\documentclass[12pt]{article}

\IfFileExists{xurl.sty}{\usepackage{xurl}}{\usepackage{url}}
\usepackage[hidelinks]{hyperref}
\usepackage[final]{microtype}
\microtypecontext{spacing=nonfrench}
\IfFileExists{doi.sty}{\usepackage{doi}}{}

\usepackage[margin=1in]{geometry}
\usepackage{setspace}
\usepackage{times}

\usepackage[utf8]{inputenc}
\usepackage[T1]{fontenc}
\usepackage[super,sort&compress]{natbib}

\usepackage{graphicx}
\usepackage{amsmath}
\usepackage{booktabs}
\usepackage{longtable}
\usepackage{array}
\IfFileExists{multirow.sty}{\usepackage{multirow}}{}
\usepackage{xcolor}
\usepackage{tabularx}
\newcolumntype{Y}{>{\raggedright\arraybackslash}X}
\usepackage[font=small,labelfont=bf]{caption}
\providecommand{\linenumbers}{}
\providecommand{\nolinenumbers}{}

\newcommand{\captionsource}[1]{\\\textit{Source information:} #1}

\title{\textbf{GOLDMARK: Governed Outcome-Linked Diagnostic Model Assessment Reference Kit}}

\author{
\parbox{\textwidth}{\centering
\small
Chad Vanderbilt\textsuperscript{1}\thanks{Corresponding author: \href{mailto:vanderbc@mskcc.org}{vanderbc@mskcc.org}},\;
Gabriele Campanella\textsuperscript{2},\;
Siddharth Singi\textsuperscript{1},\;
Swaraj Nanda\textsuperscript{1},\;
Jie\mbox{-}Fu Chen\textsuperscript{1},\;
Ali Kamali\textsuperscript{1},\;
Amir Momeni Boroujeni\textsuperscript{1},\;
David Kim\textsuperscript{1},\;
Mohamed Yakoub\textsuperscript{1},\;
Jamal Benhamida\textsuperscript{1},\;
Meera Hameed\textsuperscript{1},\;
Neeraj Kumar\textsuperscript{1,*},\;
Gregory Goldgof\textsuperscript{1,*} \\[0.5em]
\textsuperscript{*}These authors contributed equally to this work.
}
}
\date{}



\begin{document}

\nolinenumbers
{\singlespacing\maketitle}

\begin{center}
\small
\textsuperscript{1}Department of Pathology and Laboratory Medicine, Memorial Sloan Kettering Cancer Center, New York, NY, USA\\
\textsuperscript{2}Hasso Plattner Institute for Digital Health at Mount Sinai, Icahn School of Medicine at Mount Sinai, New York, NY, USA
\end{center}

\vspace{0.75em}
\noindent\textbf{Short title:} GOLDMARK

\linenumbers

\section*{Abstract}
\textbf{Background:} Computational biomarkers (CBs) are histopathology-derived patterns extracted from hematoxylin–eosin (H\&E) whole-slide images (WSIs) using artificial intelligence (AI) to predict therapeutic response or prognosis. Recently, slide-level \emph{multiple-instance learning} (MIL) with pathology foundation models (PFMs) has become the standard baseline for CB development. While these methods, with architectural and optimization advances, have improved predictive performance, computational pathology lacks standardized intermediate data formats, provenance tracking, checkpointing conventions, and reproducible evaluation metrics required for clinical-grade deployment. Consequently, discipline-level standardization, including data representation, model versioning, evaluation protocols, and auditability, is essential to enable reliable, scalable, and regulatory-ready clinical translation of CBs.

\textbf{Methods:} We introduce \textbf{GOLDMARK: Governed Outcome-Linked Diagnostic Model Assessment Reference Kit} (\emph{www.artificialintelligencepathology.org}), a standardized benchmarking framework built on a curated TCGA cohort with clinically anchored OncoKB level 1--3 biomarker labels. GOLDMARK distributes structured intermediate outputs, including tile coordinates, per-slide feature embeddings from canonical PFMs, embedding-level quality-control metadata, trained slide-level weights, and reference code. Multiple publicly available PFMs are benchmarked under a unified attention-based MIL head using predefined patient-level splits. Models are trained on TCGA and evaluated on an independent MSKCC cohort with reciprocal testing.

\textbf{Results:} We evaluated 33 tumor–biomarker tasks; aggregate summaries over the 33 tasks with complete reciprocal metric coverage yielded mean AUROC of 0.689 (TCGA) and 0.630 (MSKCC). Restricting analysis to the eight highest-performing tasks yielded mean AUROCs of 0.831 and 0.801, respectively. These tasks correspond to established morphologic–genomic associations (e.g., LGG \emph{IDH1}, COAD MSI/\emph{BRAF}, THCA \emph{BRAF}/\emph{NRAS}, BLCA \emph{FGFR3}, UCEC \emph{PTEN}) and showed the most stable cross-site performance. Differences between canonical encoders were modest relative to task-specific variability.

\textbf{Conclusions:} Computational pathology is entering a translational phase in which reproducibility, transparency, and cross-institutional robustness are prerequisites for clinical trust. GOLDMARK establishes a reference framework that separates dataset curation from model evaluation and introduces structured intermediate artifacts, quality-control metadata, and symmetric cross-dataset testing as core components of benchmarking. Such infrastructure is essential for transforming computational biomarkers from research demonstrations into reproducible, clinically trusted workflows.

\section*{Key Points}
\begin{itemize}
  \item \textbf{Practice guidance} for slide-level MIL biomarker modeling with clinically anchored labels and cross-institutional validation.
  \item \textbf{Open resources}: per-slide PFM embeddings, tile coordinate manifests, embedding metadata for quality control (QC), trained slide-level MIL weights, end-to-end reference code, case-level results, and performance-analysis visualizations.
  \item \textbf{Curation is critical}: FFPE-only TCGA selection and \textbf{OncoKB}-based labels (levels 1--3) improve label fidelity and restrict analysis to clinically actionable associations.\cite{Chakravarty2017OncoKB}
  \item \textbf{Generalization}: TCGA$\rightarrow$MSKCC (and reciprocal) testing identifies robust tasks (e.g., LGG \emph{IDH1}, CRC MSI) and common failure modes under domain shift.\cite{Campanella2019NatMed}
  \item \textbf{Interactive exploration}: target-level ROC/PR curves and an attention-enabled WSI viewer support error analysis and qualitative review of model behavior.
  \item \textbf{Sharing}: all artifacts are delivered via a public, read-only web portal with organized, versioned downloads.
\end{itemize}

\section*{Introduction}
Computational biomarkers apply contemporary artificial intelligence (AI) methods to predict treatment response or prognosis directly from routine clinical diagnostics. In surgical pathology, hematoxylin--eosin (H\&E) whole-slide images (WSIs) are the universal substrate of diagnosis, as every diagnostic case includes at least one H\&E slide. Public repositories that pair WSIs with tumor--normal sequencing have enabled morpho-genomic studies at scale by jointly sharing digital histology and matched genomic results. Among these, The Cancer Genome Atlas (TCGA) provides resection-specimen WSIs with matched tumor and \emph{normal} sequencing, enabling germline filtering and clinically meaningful somatic mutation labels. In clinical practice, targeted sequencing panels (e.g., MSK-IMPACT) are routine across many tumor types, motivating complementary approaches that leverage H\&E morphology for clinically relevant prediction.

Slide-level \emph{multiple-instance learning} (MIL) has become the dominant paradigm for WSI classification. MIL is a weakly supervised setting in which each WSI is partitioned into tiles that form a bag; a slide-level label supervises training; and an aggregator (e.g., gated MIL attention; GMA) combines tile features into slide-level predictions. In most modern pipelines, tile features are extracted with a pretrained self-supervised encoder (a pathology foundation model, PFM), typically a vision transformer trained on large corpora of unlabeled WSIs, and a lightweight MIL head is trained on top of these fixed embeddings~\cite{Chen2024UNI,Vorontsov2024Virchow,Xu2024Nature}.

GOLDMARK is not an attempt to advance the algorithmic state-of-the-art. Rather, we intentionally standardize the most widely adopted and well-trod framework for computational biomarkers: fixed feature extraction with PFMs followed by MIL aggregation. While other efforts have focused on benchmarking performance or guiding clinical deployment and validation of digital pathology tools,\cite{Campanella2025NatComm,neidlinger_benchmarking,janowczyk_guide_2025} the primary motivation for GOLDMARK is to establish interoperable intermediate processes and file standards with reproducible evaluation conventions for this canonical pipeline. Within this framework, the field can extend consistent standards to rapidly evolving paradigms, including end-to-end PFM fine-tuning and task adaptation (e.g., EAGLE, TAPFM),\cite{EAGLE2025,TAPFM2025} whole-slide self-supervised models (e.g., Prov-GigaPath, TITAN),\cite{Xu2024Nature,titan_2025} and multimodal vision--language foundation models (e.g., CONCH, MUSK)\cite{conch2024,xiang_visionlanguage_2025}.

Despite rapid advances in computational pathology, reproducibility remains hampered by heterogeneous design choices even on publicly available datasets. Decisions such as slide selection, cohort inclusion, and mutation labeling materially affect reported performance. Moreover, the absence of shared intermediate outputs—such as tile coordinate manifests, slide-level PFM embeddings, and trained MIL weights—limits reproducibility and interpretability of benchmarking studies. Feature extraction for curated TCGA cohorts with modern encoders can also require hundreds of high-memory GPU-hours, creating a substantial barrier to entry.

These limitations leave AI-based actionable biomarkers far from widespread clinical implementation. Although recent work has begun to define practical frameworks for deployment, validation, and accreditation of digital pathology tools within clinical workflows,\cite{janowczyk_guide_2025} such efforts primarily address downstream implementation and local validation. In contrast, upstream standardization of intermediate computational artifacts, quality metrics, and cross-dataset evaluation remains largely unaddressed. This situation echoes the early transition of next-generation sequencing (NGS) from a research instrument to a routine clinical test, where adoption depended not only on improved assays, but also on interoperable intermediate formats (e.g., \emph{FASTQ}, \emph{SAM/BAM}, \emph{VCF}), shared processing frameworks (e.g., \emph{samtools}, \emph{GATK}, \emph{PICARD}), and validation practices that enabled reproducibility across institutions.\cite{Li2009SAMtools,Danecek2011VCFtools,McKenna2010GATK,Picard2019toolkit} Computational pathology is now at a comparable inflection point.

We address these challenges with a standardized resource and protocol designed to improve label fidelity, comparability, and reproducible evaluation. Specifically,
\begin{itemize}
    \item we curate FFPE-only TCGA diagnostic slides and define clinically anchored mutation labels using tumor-specific OncoKB levels 1--3~\cite{Chakravarty2017OncoKB}, 
    \item release intermediate outputs (tile coordinate manifests, per-slide PFM embeddings with lightweight QA/QC metadata, and baseline slide-level MIL weights),
    \item evaluate generalization with TCGA$\rightarrow$MSKCC testing plus reciprocal MSKCC$\rightarrow$TCGA experiments under identical preprocessing, and
    \item provide GOLDMARK as a public, open-source platform (\url{https://www.artificialintelligencepathology.org}) for interactive exploration, including downloadable artifacts, model outputs, and attention-based slide visualization.
\end{itemize}

\begin{figure}[t]
\centering
\includegraphics[width=0.95\textwidth]{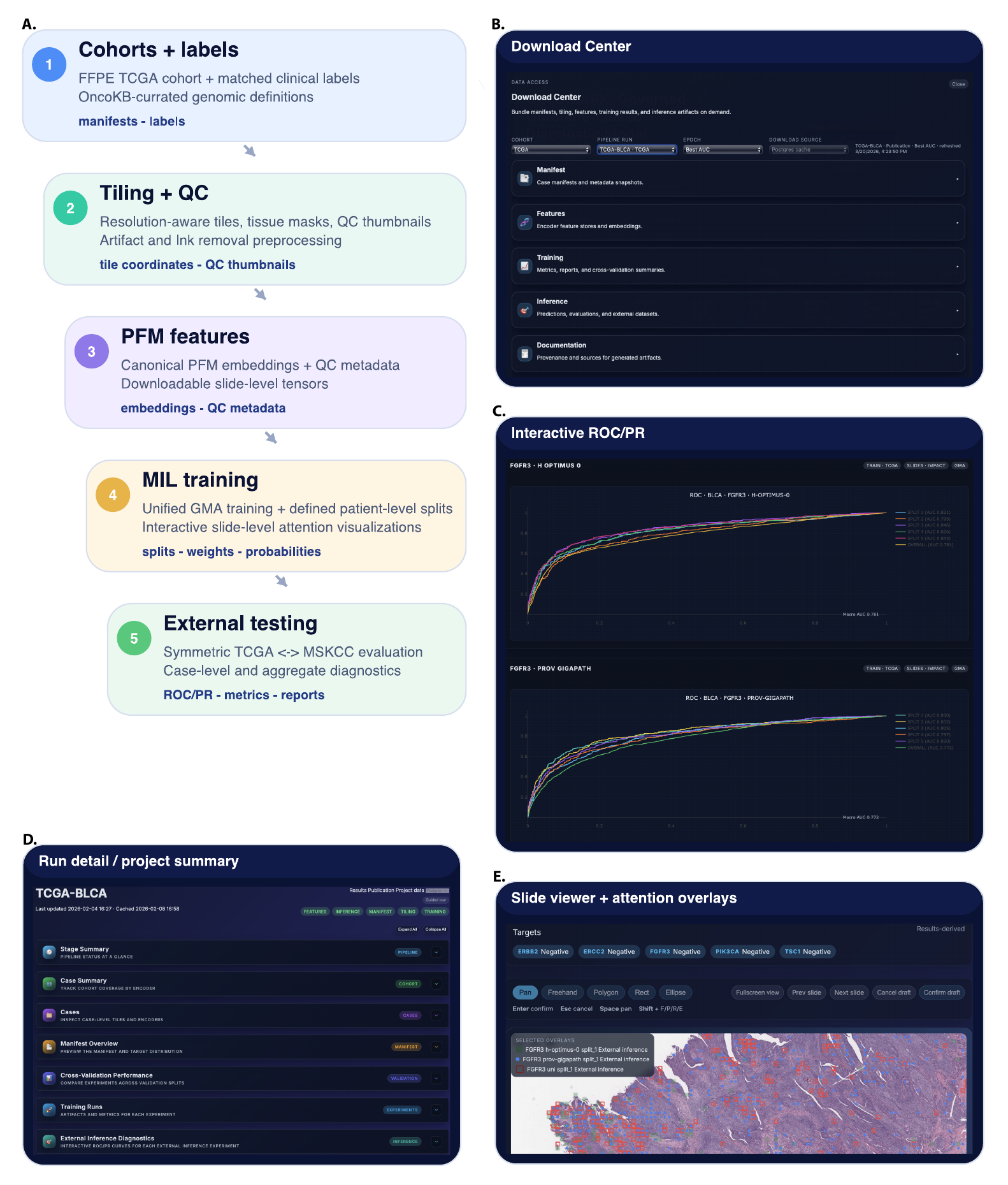}
\vspace{-0.5em}
\caption{\textbf{GOLDMARK workflow and \href{https://artificialintelligencepathology.org}{artificialintelligencepathology.org}.} (A) Standardized pipeline from curated TCGA cohorts and OncoKB labels through tiling, PFM feature extraction, MIL training with predefined splits, and symmetric TCGA$\leftrightarrow$MSKCC evaluation, producing structured intermediate artifacts (manifests, tile coordinate manifests, embeddings, splits, weights, and metrics). (B--E) Public website screenshots: (B) Download Center for intermediate files with exposed application programming interfaces (APIs), (C) interactive ROC/PR performance plots, (D) run-level summaries and experiment tracking, and (E) slide viewer with interactive overlay visualizations of model attention.\captionsource{Pipeline schematic, portal screenshots, and associated analyses from GOLDMARK.}}
\vspace{-0.75em}
\label{fig:workflow}
\end{figure}

GOLDMARK includes models built from multiple PFMs with attention-based MIL using patient-level splits, provides per-task performance for TCGA cross-validation and external MSKCC testing, and distills practical recommendations for curation, bag design, sampling, labeling, and validation. By releasing per-slide embeddings, tile coordinate manifests, predefined splits, and baseline slide-level weights, GOLDMARK establishes a shared experimental substrate that enables \emph{direct, controlled comparison of methods} and promotes transparent, reproducible experimentation.

\section*{Methods}

This study was approved by the Memorial Sloan Kettering Cancer Center (MSKCC) Institutional Review Board (IRB 18-013). Non-public TCGA materials were accessed via the Database of Genotypes and Phenotypes (dbGaP) under an approved data-access protocol in accordance with National Institutes of Health data-use policies.

In addition to GOLDMARK (\texttt{www.artificialintelligencepathology.org}), which displays and releases data in a user-friendly interface, the curation, preprocessing, training, and inference workflows used in this study are publicly available through a GitHub repository (\url{https://github.com/chadvanderbilt/GOLDMARK}). The pipeline is constructed such that appropriate slides are curated automatically with genomic targets assigned through Genomic Data Commons (GDC)-based downloads and automated OncoKB API labeling. All model training and inference steps described below are fully implemented, including resolution- and tissue-aware tile extraction, pathology foundation model (PFM) feature extraction with quality-control safeguards, standardized split generation, slide-level GMA model training, and internal and reciprocal external inference.

\subsection*{Cohorts and Curation}

\paragraph{TCGA (training and cross-validation).}
We assemble diagnostic H\&E WSIs from TCGA and restrict to \textbf{formalin-fixed paraffin-embedded (FFPE)} tissue to avoid frozen-section artifacts and to anchor training to the most common preparation received for genomics testing. Slides are identified programmatically using GDC metadata and file naming, prioritizing primary diagnostic material (filenames containing ``\mbox{-00-DX}''). Slides without matched tumor--normal sequencing data or with frozen preparation are excluded.\cite{GDCAPI}
Mutation tasks are defined where OncoKB designates the tumor-variant pairing as clinically actionable (levels 1--3), ensuring labels align to current clinical use.\cite{Chakravarty2017OncoKB} Point mutations and small indels are considered. \emph{POLE} mutations are prognostically relevant but not associated with treatment; \emph{POLE}-positive uterine carcinoma is curated based on hotspot mutations.\cite{nero_pole_2025} Microsatellite instability is assigned using consensus of MSISensor and MiMSI with standard settings.\cite{niu_msisensor_2014,Ziegler2025MiMSI}

\paragraph{MSKCC (external testing).}
The external clinical cohort comprises H\&E WSIs scanned from the same tissue blocks tested by \textbf{MSK-IMPACT} under routine care; ground-truth alterations are obtained from clinical reports generated by board-certified molecular pathologists.\cite{Cheng2015JMolDiagn,Zehir2017NatMed} Tumor types are aligned by pairing OncoTree codes assigned at diagnosis with TCGA project names. Mutations are curated using an identical pipeline to assign OncoKB labels and provide binary labels for each biomarker. We perform the reciprocal experiment (MSKCC$\rightarrow$TCGA) under identical definitions.

\subsection*{Slide Selection and Quality Control}

Non-H\&E stains and frozen section material are excluded to ensure analysis is restricted to diagnostic H\&E slides. TCGA images were acquired using Aperio scanners at 20$\times$ or 40$\times$ objective-equivalent magnification, while MSKCC slides were digitized using a combination of Aperio AT2 (20$\times$) and GT450 (40$\times$) scanners. To address scanner and magnification variability, images are harmonized based on \emph{physical resolution} (microns-per-pixel, mpp) rather than nominal magnification.\cite{goode_openslide_2013}
Slides with insufficient tissue area (below 25~mm$^2$ total tissue after segmentation), gross scanning artifacts (large folds, out-of-focus throughout), corrupted \texttt{.svs} files that fail to open with standard tools, or missing/implausible mpp metadata are excluded. When multiple duplicate scans existed for the same glass slide or case, the highest-resolution, least compressed image was retained. The general workflow is documented in Figure~\ref{fig:workflow}A. 

\subsection*{Image Tiling}

Whole-slide images are divided into small, square \emph{tiles} covering tissue regions, with all steps performed in a resolution-aware manner to ensure comparability across scanners. We standardize the physical field-of-view to \textbf{128~$\mu$m} per tile; for a slide with microns-per-pixel $m$ ($\mu$m/px), the tile width in pixels is computed as
\[
\text{tile\_px} = \mathrm{round}\left(\frac{128}{m}\right),
\]
yielding, for example, 256 pixels at 0.50~$\mu$m/px (typical 20$\times$) and 512 pixels at 0.25~$\mu$m/px (typical 40$\times$), thereby preserving a constant biological field-of-view. Tiles are placed on a regular grid over tissue with a \textbf{1 pixel} non-overlap to reduce boundary effects and improve coverage of small foci, and all tiles meeting inclusion criteria are retained for downstream analysis. Tissue detection is performed on a low-resolution thumbnail generated with OpenSlide: the image is converted to grayscale, smoothed with a Gaussian filter, segmented using Otsu thresholding to separate tissue from background, followed by suppression of low-variance background regions and exclusion of pen/marker artifacts via color-based heuristics. For each accepted tile, the pipeline records top-left pixel coordinates in native slide space together with slide and target identifiers and tiling metadata (e.g., fraction tissue, pixel size) to enable reproducible downstream extraction of tiles. Separate tile coordinate files for 20$\times$ and 40$\times$ slides are downloadable for each TCGA project. For each WSI, we additionally generate a per-slide tile manifest (CSV) to be used for downstream attention overlays and tile--feature integrity checks. For all slides in the study, thumbnails with tile-grid overlays are viewable via the GOLDMARK portal for visual quality control by investigators and pathologists.

\subsection*{Feature Extraction with Pathology Foundation Models}

We computed numeric feature vectors for each tile using representative open PFMs (e.g., UNI, Virchow/Virchow2, Prov-GigaPath, EAGLE (aka gigapath\_ft), and H-Optimus-0).\cite{Chen2024UNI,Vorontsov2024Virchow,Xu2024Nature,EAGLE2025} Encoders were used strictly as fixed feature extractors. For each slide, all tile embeddings are concatenated into a single per-slide feature file while preserving the exact one-to-one mapping to tile coordinates, enabling deterministic reconstruction of tile-feature relationships for attention overlays in whole slide image viewer. Because this step is compute-intensive, we release the per-slide feature files to allow downstream slide-level modeling without repeating extraction, downloadable via GOLDMARK API. Each per-slide feature file additionally includes summary statistics (e.g., \texttt{embedding\_variance}, feature norm ranges, NaN/Inf fractions, near-duplicate estimates) to enable rapid sanity checks, cross-encoder comparisons, and detection of extraction failures such as truncated tensors, thereby informing downstream normalization decisions. Finally, each feature tensor is paired with a metadata JSON file that records extraction provenance and a checksum to support integrity verification across storage and transfer.

\subsection*{Genomic Label Definition (OncoKB-Anchored)}

Alteration labels are derived from mutation annotation files (MAFs) downloaded using GDC and paired with the digital slide based on the case identifier. All annotated variants represent calls that are somatic, meaning the variants are present in the tumor but absent from the normal. Only mutations in cancer genes (defined by OncoKB) are evaluated for oncogenic status. Each mutation, based on genomic coordinates and the observed alteration from reference, is mapped to clinical actionability using \textbf{OncoKB} levels 1--3 \emph{for the specific tumor type}. Variants of uncertain significance are excluded. OncoKB actionability is resolved programmatically via API lookups. This produces binary labels (present/absent) for each tumor--gene task. Mutation-specific tasks are included if TCGA has at least 15 positive cases for a given target.

\subsection*{Slide-Level Learning (Multiple-Instance Learning)}

Each slide is modeled as a bag of tile features. We train a gated-attention MIL model that learns to (i) assign weights to tiles according to relevance and (ii) combine them into slide predictions.\cite{Ilse2018ABMIL,Lu2021CLAM} Features are fixed; only the slide-level head is trained. The GOLDMARK portal allows downloading of trained slide-level model parameters via API.

\subsection*{Training, Splits, External Testing, and Evaluation}

We perform five stratified 70/30 patient-level splits within TCGA for cross-validation, preserving class balance and preventing patient overlap between training and testing folds. Models are trained for 120 epochs per split using fixed hyperparameters across all tasks to avoid biomarker-specific overfitting, and split assignments are persisted in a versioned manifest and reused across encoders and evaluation contexts to ensure identical patient partitions and fair comparisons across experiments. Model selection is based on cross-validation performance within TCGA, with the best-performing epoch identified per split; inference is conducted using both (i) the best-AUC epoch and (ii) the final trained (120th) epoch checkpoint for consistency of reported results. After training on each TCGA split, slide-level models are frozen and applied to the MSKCC IMPACT cohort, performing inference on every available slide, and we additionally conduct the symmetric analysis (MSKCC$\rightarrow$TCGA) under identical preprocessing and training procedures to characterize bidirectional generalization. Discrimination performance is evaluated primarily using area under the receiver-operating-characteristic curve (AUROC), with precision--recall curves provided on the GOLDMARK portal; variability and confidence intervals are summarized across the five repeated splits to capture uncertainty under independent random partitions. Probability calibration is assessed using binned reliability diagrams (predicted vs. observed event rates) computed from model probabilities, and calibration plots are available for download to quantify agreement between predicted probabilities and observed event frequencies.\cite{Guo2017Calibration}

\subsection*{GOLDMARK Portal: Interactive Exploration and Downloads}
\label{sec:website}

GOLDMARK provides a public web portal (\emph{www.artificialintelligencepathology.org}) for interactive exploration of results and access to downloadable artifacts. The portal operates as a public deployment and serves precomputed outputs without requiring specialized hardware. Views from the the GOLDMARK portal are shown in Figure~\ref{fig:workflow}B-E. It provides organized, versioned bundles for manifests, tiling coordinates, per-slide PFM embeddings, slide-level model weights for all trained models (TCGA and MSKCC), and per-slide evaluation scores, alongside interactive visualizations of cross-validation and cross-dataset inference performance. An integrated whole-slide viewer enables overlay of per-tile attention maps and annotations, with tools to highlight top-percentile attention regions to support qualitative review and error analysis. For each held-out test fold, per-tile attention scores can be exported as CSV files aligned to slide-level tile manifests, enabling deterministic reconstruction of overlays outside the website; attention outputs are saved for both the validation-selected best checkpoint and a fixed late training epoch (e.g., epoch 120).

\subsection*{Integrity checks (tiling \texorpdfstring{$\leftrightarrow$}{<->} features)}
\label{sec:integrity}

To support quality assurance at scale, we (i) verify \textbf{tile--feature count consistency} for each slide and encoder by checking that the feature tensor length equals the number of tiles listed in the tile coordinate manifest (discrepancies flag partial feature extraction or mismatched manifests) and (ii) provide \textbf{coverage visualization} via cached thumbnails with tiling overlays and point-cloud views of coordinate maps to verify that extracted tiles cover tissue regions. Artifacts failing tile--feature cardinality checks are explicitly marked as failed (e.g., with a failure suffix) and excluded from downstream manifests. Pipeline execution is fail-closed: training and inference proceed only when required slide-level feature artifacts are present, pass quality checks, and match the corresponding tile manifests. We additionally flag degenerate extractions using low embedding-variance thresholds, defined as per-slide feature tensors exhibiting near-constant values across tiles. Such low variance may indicate feature-extraction failures, an encoder that does not meaningfully differentiate morphologic patterns, or slides with intrinsically limited morphologic diversity (e.g., diffuse blood or extensive necrosis), as well as technical artifacts related to tissue preparation, scanning, or improper code implementation. Slides meeting this criterion are flagged for quality control and are excluded from downstream training or inference analyses.

\section*{Results}

\subsection*{Benchmark Overview}
We evaluated 33 tumor type--biomarker tasks spanning 14 tumor types with matched cohorts in TCGA and MSKCC (clinical WSIs paired to MSK-IMPACT testing). Task definitions were restricted to clinically actionable alterations by anchoring labels to tumor-specific \textbf{OncoKB evidence levels 1--3}.\cite{Chakravarty2017OncoKB} Table~\ref{tab:gene_mutation_counts} summarizes paired case counts and prevalences, illustrating the shift from TCGA's modest public cohorts to MSKCC's higher-volume, real-world clinical testing population. Aggregate performance summaries below focus on the 33 tasks with complete reciprocal metric coverage across PFM encoders and both dataset directions.

\begin{table}[ht]
\centering
\resizebox{\textwidth}{!}{%
\begin{tabular}{l l l r r r r r r}
\toprule
Tumor Code & Tumor Type & Gene Mutation &
\multicolumn{3}{c}{MSKCC} &
\multicolumn{3}{c}{TCGA} \\
\cmidrule(lr){4-6} \cmidrule(lr){7-9}
 &  &  & Total & Positive & Negative & Total & Positive & Negative \\
\midrule
BLCA & Bladder Urothelial Carcinoma & PIK3CA & 2031 & 337 & 1694 & 386 & 77 & 309 \\
BLCA & Bladder Urothelial Carcinoma & FGFR3 & 2031 & 393 & 1638 & 386 & 34 & 352 \\
BLCA & Bladder Urothelial Carcinoma & ERBB2 & 2031 & 207 & 1824 & 386 & 37 & 349 \\
BLCA & Bladder Urothelial Carcinoma & TSC1 & 2031 & 131 & 1900 & 386 & 27 & 359 \\
BLCA & Bladder Urothelial Carcinoma & ERCC2 & 2031 & 141 & 1890 & 386 & 28 & 358 \\
BRCA & Breast Carcinoma & PIK3CA & 2735 & 966 & 1769 & 1000 & 354 & 646 \\
CESC & Cervical Squamous Cell Carcinoma & PIK3CA & 171 & 62 & 109 & 266 & 82 & 184 \\
COAD & Colon Adenocarcinoma & PIK3CA & 2959 & 605 & 2354 & 550 & 130 & 420 \\
COAD & Colon Adenocarcinoma & BRAF & 2959 & 318 & 2641 & 550 & 59 & 491 \\
COAD & Colon Adenocarcinoma & ATM & 2959 & 153 & 2806 & 550 & 63 & 487 \\
COAD & Colon Adenocarcinoma & PTEN & 2959 & 183 & 2776 & 550 & 38 & 512 \\
COAD & Colon Adenocarcinoma & MSI & 2959 & 350 & 2609 & 405 & 70 & 335 \\
GBM & Glioblastoma Multiforme & PTEN & 816 & 263 & 553 & 244 & 77 & 167 \\
UCEC & Endometrial Cancer & PTEN & 2062 & 972 & 1090 & 499 & 319 & 180 \\
UCEC & Endometrial Cancer & PIK3CA & 2062 & 904 & 1158 & 499 & 254 & 245 \\
UCEC & Endometrial Cancer & FBXW7 & 2062 & 279 & 1783 & 499 & 94 & 405 \\
UCEC & Endometrial Cancer & ATM & 2062 & 142 & 1920 & 499 & 73 & 426 \\
UCEC & Endometrial Cancer & POLE & 2062 & 114 & 1948 & 499 & 49 & 450 \\
LGG & Glioma & IDH1 & 449 & 216 & 233 & 491 & 382 & 109 \\
LGG & Glioma & PIK3CA & 449 & 41 & 408 & 491 & 42 & 449 \\
HNSC & Head and Neck Carcinoma & PIK3CA & 485 & 84 & 401 & 431 & 66 & 365 \\
HNSC & Head and Neck Carcinoma & HRAS & 485 & 11 & 474 & 431 & 25 & 406 \\
LUAD & Lung Adenocarcinoma & KRAS & 923 & 273 & 650 & 465 & 171 & 294 \\
LUAD & Lung Adenocarcinoma & EGFR & 923 & 273 & 650 & 465 & 51 & 414 \\
SKCM & Melanoma & BRAF & 886 & 213 & 673 & 432 & 233 & 199 \\
SKCM & Melanoma & NRAS & 886 & 143 & 743 & 432 & 112 & 320 \\
SKCM & Melanoma & PTEN & 886 & 63 & 823 & 432 & 50 & 382 \\
SKCM & Melanoma & MAP2K1 & 886 & 38 & 848 & 432 & 26 & 406 \\
PAAD & Pancreatic Adenocarcinoma & KRAS & 3018 & 2674 & 344 & 181 & 136 & 45 \\
PCPG & Pheochromocytoma/Paraganglioma & HRAS & 18 & 2 & 16 & 176 & 19 & 157 \\
STAD & Stomach Adenocarcinoma & PIK3CA & 742 & 69 & 673 & 374 & 60 & 314 \\
THCA & Thyroid Cancer & BRAF & 920 & 399 & 521 & 495 & 296 & 199 \\
THCA & Thyroid Cancer & NRAS & 920 & 130 & 790 & 495 & 41 & 454 \\
\bottomrule
\end{tabular}%
}
\caption{Gene mutation counts by tumor type (MSKCC and TCGA).}
\label{tab:gene_mutation_counts}
\end{table}

All models were trained with patient-level splits and evaluated under four complementary settings: (i) TCGA$\rightarrow$TCGA cross-validation, (ii) TCGA$\rightarrow$MSKCC external testing, and the reciprocal direction, (iii) MSKCC$\rightarrow$MSKCC cross-validation and (iv) MSKCC$\rightarrow$TCGA external testing. For external testing, slide-level models were applied with \emph{frozen weights} (no recalibration, no re-sampling, no fine-tuning), so any performance difference reflects generalization under deployment-like conditions.

\subsection*{Task Performance}
\begin{figure}[t]
\centering
\includegraphics[width=0.95\textwidth]{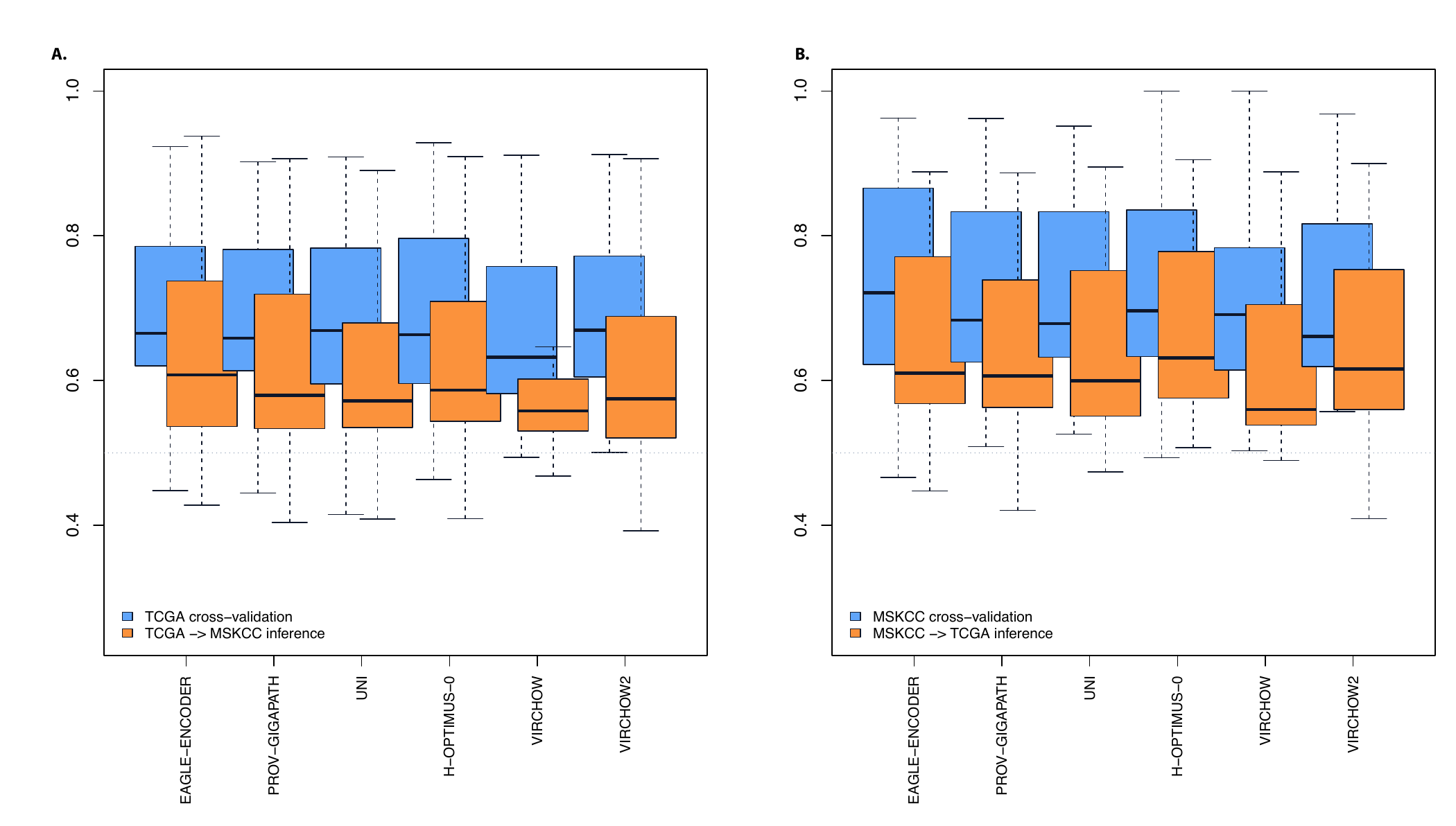}
\vspace{-0.5em}
\caption{\textbf{Cross-validation and reciprocal external testing across tumor–biomarker tasks and encoders.} (A) TCGA cross-validation compared with frozen-weight inference on MSKCC. (B) MSKCC cross-validation compared with frozen-weight inference on TCGA. Boxplots summarize AUROC distributions across tumor–biomarker tasks using repeated patient-level splits.\captionsource{GOLDMARK analyses of curated TCGA and MSKCC cohorts.}}
\vspace{-0.75em}
\label{fig:performance_multipanel}
\end{figure}

Figure~\ref{fig:performance_multipanel} summarizes AUROC distributions across the 33 tasks with complete reciprocal metric coverage and across encoders (UNI, Virchow/Virchow2, Prov-GigaPath, and h-optimus-0, plus the task-adapted EAGLE encoder). In aggregate, mean AUROC decreased from 0.689 in TCGA cross-validation to 0.630 in MSKCC external testing, consistent with expected domain shift in clinical slide data (scanner, staining, case mix, and sampling). The reciprocal evaluation revealed a similar pattern: tasks that generalized well from TCGA$\rightarrow$MSKCC frequently generalized well in the reverse direction, whereas many weaker tasks remained near-chance regardless of direction, reflecting poor correlations between genomic result and a detectable morphology.

Across tasks, the dominant source of variation was \emph{task difficulty}, not encoder choice. For most biomarkers, AUROC distributions across PFM encoders were highly overlapping, indicating that with fixed features and an attention-based MIL head encoder choice typically yields incremental rather than dramatic gains. The dramatic exception is that the EAGLE encoder is adapted specifically for the detection of \emph{EGFR} in lung adenocarcinoma. Even though this experiment lacks the co-adapted aggregation function, the encoder alone markedly outperformed other encoders on this task across all experimental settings.

The general lack of performance differentiation across encoders motivates treating PFM encoder selection as an important but secondary decision relative to cohort curation and clinically anchored label definitions. In practical terms, GOLDMARK's primary utility is enabling fair comparisons where the largest sources of variability (data selection, labeling, and evaluation) are standardized and logged.

To summarize encoder ordering without over-interpreting small absolute differences, we ranked encoders \emph{within each task} by AUROC and aggregated ranks across tasks (Borda-style rank-sum; lower is better). Table~\ref{tab:encoder_ranking} shows that \texttt{EAGLE} (a fine-tuned version of \texttt{prov-gigapath}) and \texttt{h-optimus-0} consistently ranked highest overall, with \texttt{virchow2} and \texttt{prov-gigapath} close behind in both TCGA cross-validation and TCGA$\rightarrow$MSKCC external testing. Nevertheless, absolute mean AUROC differences between encoders were modest, reinforcing that robust external validation, standardized preprocessing, and task selection grounded in clinical morphology are at least as important as the specific choice of canonical encoder.

\begin{table}[t]
\centering
\small
\caption{\textbf{Encoder performance and rank aggregation across tasks.} For each task, encoders are ranked by AUROC and ranks are summed across tasks (lower rank-sum indicates better overall performance). Top: all evaluated tasks with complete reciprocal metric coverage. Bottom: the top eight tasks selected by combined median AUROC across encoders and evaluation contexts (Figure~\ref{fig:top8_grid}). Bold text indicates the best model in each setting.\captionsource{GOLDMARK analyses of curated TCGA and MSKCC cohorts.}}
\label{tab:encoder_ranking}
\begin{tabular}{lcccc}
\toprule
\multicolumn{5}{c}{All tasks (complete reciprocal coverage subset; $n=33$)}\\
\midrule
Encoder
& \shortstack{Mean AUROC\\(TCGA-CV)}
& Rank-sum
& \shortstack{Mean AUROC\\(TCGA$\rightarrow$MSKCC)}
& Rank-sum \\
\midrule
\texttt{EAGLE} & \textbf{0.706} & \textbf{76}  & \textbf{0.647} & \textbf{88}  \\
\texttt{h-optimus-0}         & 0.692 & 98  & 0.639 & 84  \\
\texttt{virchow2}            & 0.690 & 104 & 0.630 & 114 \\
\texttt{prov-gigapath}       & 0.694 & 108 & 0.635 & 111 \\
\texttt{uni}                 & 0.682 & 127 & 0.625 & 117 \\
\texttt{virchow}             & 0.671 & 138 & 0.606 & 137 \\
\midrule
\multicolumn{5}{c}{Top eight tasks (selected by combined median AUROC)}\\
\midrule
Encoder
& \shortstack{Mean AUROC\\(TCGA-CV)}
& Rank-sum
& \shortstack{Mean AUROC\\(TCGA$\rightarrow$MSKCC)}
& Rank-sum \\
\midrule
\texttt{h-optimus-0}         & \textbf{0.850} & \textbf{14} & 0.810 & 23 \\
\texttt{EAGLE}               & 0.855 & 16 & \textbf{0.825} & \textbf{21} \\
\texttt{virchow2}            & 0.824 & 30 & 0.805 & 24 \\
\texttt{prov-gigapath}       & 0.826 & 34 & 0.805 & 33 \\
\texttt{uni}                 & 0.820 & 36 & 0.794 & 29 \\
\texttt{virchow}             & 0.813 & 38 & 0.769 & 38 \\
\bottomrule
\end{tabular}
\end{table}

\subsection*{Stable Morphogenomic Signals}
\begin{figure}[t]
\centering
\includegraphics[width=0.95\textwidth]{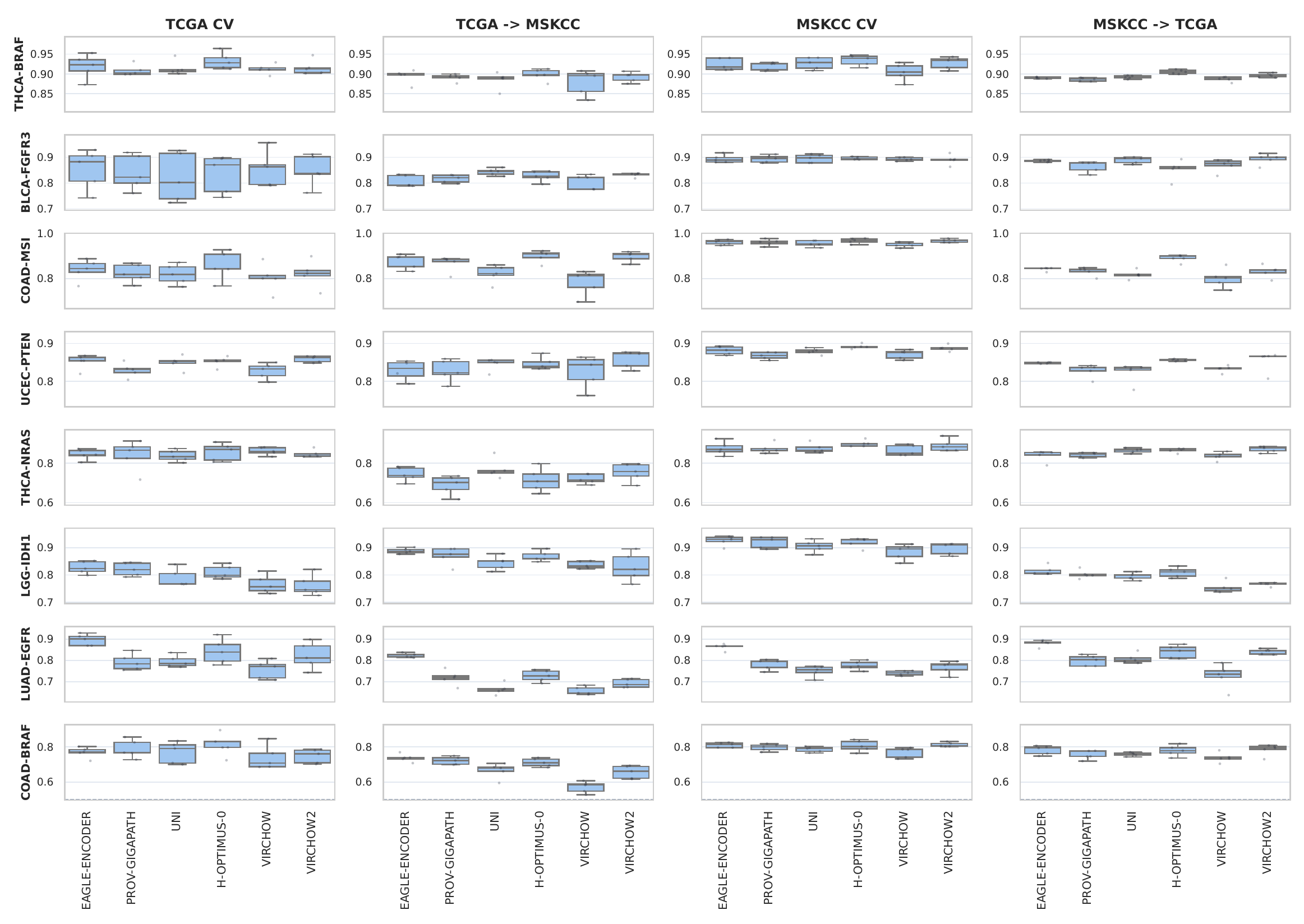}
\vspace{-0.5em}
\caption{\textbf{Top eight tumor–biomarker tasks across encoders and evaluation contexts.} Each row corresponds to a task and each column to an evaluation setting: TCGA cross-validation, TCGA$\rightarrow$MSKCC inference, MSKCC cross-validation, and MSKCC$\rightarrow$TCGA inference. Boxplots summarize AUROC distributions across five patient-level splits. Tasks were selected by combined median AUROC across encoders and evaluation contexts.\captionsource{GOLDMARK analyses of curated TCGA and MSKCC cohorts.}}
\vspace{-0.75em}
\label{fig:top8_grid}
\end{figure}

Because many biomarker tasks are difficult and may not be distinguishable from chance at current data scales or inherently impracticable, we focused interpretive analysis on the eight highest-performing tasks by combined median AUROC across encoders and reciprocal evaluation contexts (Figure~\ref{fig:top8_grid}): THCA:\emph{BRAF}, BLCA:\emph{FGFR3}, COAD:MSI, UCEC:\emph{PTEN}, THCA:\emph{NRAS}, LGG:\emph{IDH1}, LUAD:\emph{EGFR}, and COAD:\emph{BRAF}. These tasks achieved high mean performance (top-8 mean AUROC 0.831 in TCGA-CV and 0.801 in TCGA$\rightarrow$MSKCC external testing) and represent clinically actionable alterations with established or plausible morphologic correlates.

Top-task generalization was not uniformly worse in external testing. When averaged across canonical encoders, LGG:\emph{IDH1} ($\Delta$AUROC $+0.069$) and COAD:MSI ($+0.046$) improved in TCGA$\rightarrow$MSKCC testing relative to TCGA-CV, whereas THCA:\emph{NRAS} ($-0.122$), LUAD:\emph{EGFR} ($-0.105$), and COAD:\emph{BRAF} ($-0.083$) showed the largest drops. These direction-specific shifts suggest that ``domain shift'' is not purely random noise but can reflect cohort composition and practice-pattern differences (e.g., case mix, sampling strategies, and prevalence of morphologic subtypes).

\textbf{Thyroid cancer (\emph{BRAF}, \emph{NRAS}).} \emph{BRAF} in thyroid carcinoma achieved the strongest overall performance and remained robust externally. This aligns with prior work describing the association between \emph{BRAF} status and papillary thyroid carcinoma morphology and subtype composition (i.e., tall-cell variant).\cite{Lee2007BRAFThyroid} \emph{NRAS} also performed strongly in TCGA but generalized less well, consistent with a weaker or more heterogeneous morphologic phenotype relative to \emph{BRAF}-driven papillary tumors.

\textbf{Colorectal cancer (MSI, \emph{BRAF}).} MSI in colorectal cancer is associated with characteristic histologic patterns (e.g., poor differentiation and prominent lymphocytic infiltrates) that allow pathologists to prioritize cases for molecular testing.\cite{Alexander2001MSIHistopath} The stability of MSI performance across institutions is consistent with these coarse correlates and with prior computational pathology results.\cite{Kather2019NatMed} COAD:\emph{BRAF} also ranked among the top tasks, consistent with known clinicopathologic substructure in the serrated/\emph{BRAF}-mutant pathway.

\textbf{Endometrial cancer \emph{PTEN}.} \emph{PTEN} alterations are highly prevalent in endometrial cancer and are enriched in endometrioid tumors, supporting a coarse morphologic correlate that can translate across cohorts. In our benchmarking, UCEC:\emph{PTEN} achieved strong AUROC with minimal external delta, consistent with a stable morphogenomic signal under the GOLDMARK evaluation protocol.

\textbf{Bladder cancer \emph{FGFR3}.} \emph{FGFR3} alterations are enriched in papillary, low-grade urothelial carcinomas and have been linked to distinct morphologic phenotypes.\cite{alahmadie_somatic_2011,Knowles2007FGFR3} In our benchmarking, \emph{FGFR3} showed strong and often improved external performance, suggesting a robust morphologic signal under the GOLDMARK preprocessing and evaluation protocol.

\textbf{Lower-grade glioma \emph{IDH1}.} \emph{IDH1} status stratifies gliomas with well-described histopathologic and molecular differences, and its predictability from H\&E has been repeatedly demonstrated.\cite{Yan2009IDH,Coudray2018NatMed} We observed high AUROC with small cross-site deltas, supporting \emph{IDH1} as a prototypical example of a stable morphogenomic biomarker.

\textbf{Additional high-performing tasks (\emph{EGFR}, \emph{HRAS}).} LUAD:\emph{EGFR} achieved strong internal performance but showed a larger external drop on average than MSI, LGG:\emph{IDH1}, or UCEC:\emph{PTEN}, reinforcing that cross-institutional testing is needed to distinguish robust morphologic signals from cohort-sensitive effects. Campanella et al. demonstrated that an end-to-end PFM fine-tuning strategy is required to deliver strong generalization from one cohort to another.\cite{EAGLE2025,rakaee_ancestry-associated_2026} Several \emph{HRAS} tasks (e.g., PCPG:\emph{HRAS}, HNSC:\emph{HRAS}) were also among the strongest performers but exhibited more variable external behavior across cohorts.

Overall, these results reinforce a practical lens for translation: \textbf{tasks that reflect strong, coarse morphologic correlates of genotype} tend to generalize best across institutions, while tasks with weaker or heterogeneous correlates require larger cohorts, stronger supervision, or explicit domain adaptation.

\subsection*{Quality Assurance}
To support reproducibility and to enable benchmarking without re-running feature extraction, GOLDMARK releases intermediate files and associated QC signals. Across encoders, we verified \textbf{tile--feature count consistency} by requiring that each per-slide feature tensor length match the number of tiles in the coordinate manifest. On the largest evaluated cohort (COAD), tile--feature consistency was 100\% for TCGA (434/434 slides for each encoder) and 99.97\% for MSKCC (2952/2953 slides for each encoder), with a single slide flagged as having truncated feature output across encoders. This QC step is critical because partial feature extraction can silently bias analyses by effectively using only a subset of the slide.

\begin{figure}[t]
\centering
\includegraphics[width=0.95\textwidth]{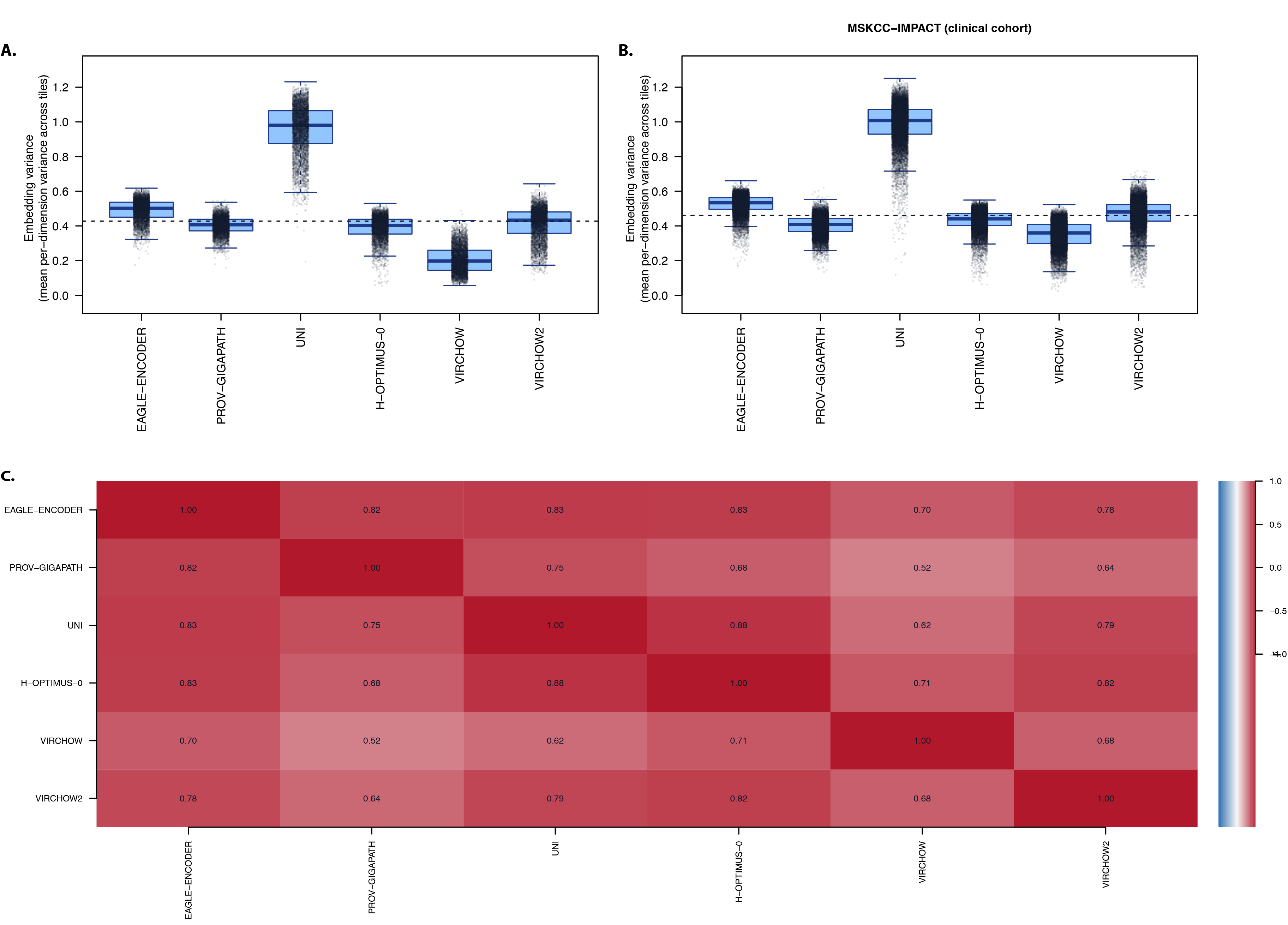}
\vspace{-0.5em}
\caption{\textbf{Embedding variance metadata across canonical encoders.} (A) TCGA cohort. (B) MSKCC cohort. Boxplots show per-slide embedding variance across encoders. (C) Cross-encoder Pearson correlations of embedding variance across slides with complete encoder coverage.\captionsource{GOLDMARK per-slide embedding metadata from curated TCGA and MSKCC cohorts.}}
\vspace{-0.75em}
\label{fig:embedding_variance}
\end{figure}

We additionally computed encoder-specific \emph{embedding metadata}, including per-slide embedding variance, as meaningful QA/QC signals that can be inspected without loading full feature tensors. Figure~\ref{fig:embedding_variance} shows that embedding variance distributions differ in scale across encoders but preserve a consistent rank-ordering across slides with strong cross-encoder correlations. These metadata provide practical safeguards for (i) detecting degenerate or truncated feature extractions, (ii) monitoring unexpected distribution shift between cohorts, and (iii) informing downstream normalization choices when benchmarking MIL heads.

Finally, we observed that training dynamics matter even for a fixed MIL head. In BLCA:\emph{FGFR3} (TCGA training; h-optimus-0 encoder), validation-selected checkpoints often occurred within the first 2--10 epochs across splits. Selecting the best checkpoint improved TCGA$\rightarrow$MSKCC AUROC by a mean of 0.039 compared with a fixed late epoch (120) across five splits (Supplementary Figure~S1). This highlights why standardized training logs and checkpoint-level metric reporting are essential for transparent, comparable benchmarking.

\subsection*{Guidance}
\label{sec:guidance}

Here we outline recommended practices for multiple-instance-learning (MIL)–based biomarker modeling in computational pathology. When training with TCGA, we recommend restricting analyses to \textbf{FFPE-only} whole-slide images (WSIs) to avoid frozen-section artifacts and color variability that can degrade performance and confound cross-study comparisons, and defining genomic mutation positivity using \textbf{tumor-specific OncoKB evidence levels 1--3}, excluding variants of uncertain significance and non-actionable associations to improve label fidelity.\cite{Chakravarty2017OncoKB} Full tiling specifications should be reported (objective equivalence, $\mu$m/px, tile size and stride, tiles per slide), with consistent physical resolution across cohorts; because the source of morphologic signal cannot be known \emph{a priori}, all tile features should be included when training slide-level models, and artifact filtering (pen marks, blur, tissue edges) should be consistently applied. For feature extraction, open pathology foundation models (PFMs) such as UNI, Virchow/Virchow2, and Prov-GigaPath may be used as fixed encoders to establish reproducible baselines,\cite{Chen2024UNI,Vorontsov2024Virchow,Xu2024Nature} training only the slide-level MIL head to enable fair aggregator comparisons and avoid encoder-specific confounders. Per-slide embedding QA/QC metadata (e.g., embedding variance, norms, NaN/Inf fractions) should be persisted to facilitate cross-encoder comparability and rapid failure detection. For aggregation and training, gated-attention MIL provides a strong, interpretable baseline with attention overlays for error analysis.\cite{Ilse2018ABMIL,Lu2021CLAM} Patient-level splits are essential to avoid leakage, with label stratification where feasible. Finally, validation should include cross-institutional testing and, when possible, reciprocal experiments to symmetrically assess generalization,\cite{Campanella2019NatMed} alongside automated QC (tile--feature count consistency, embedding monitoring) paired with lightweight visual review via tiling and attention overlays; for transparency and reuse, adherence to TRIPOD+AI is recommended, with release of manifests, tile coordinates, embeddings, and trained slide-level weights.\cite{Collins2024TRIPODAI}

\section*{Discussion}

\subsection*{AI in Histopathology}
A useful historical parallel for the current state of computational pathology is the translation of next-generation sequencing (NGS) in genomics from a research technology into a routine clinical diagnostic platform. In the early 2010s, ``genomics-driven oncology'' was motivating precision therapy selection, but widespread clinical adoption required robust, standardized, and auditable pipelines.\cite{Garraway2013GenomicsDrivenOncology} A key enabling factor was the emergence of common, interoperable file formats (e.g., FASTQ, SAM/BAM, VCF, MAF) and widely shared processing frameworks (e.g., GATK) that made analyses reproducible and failures debuggable across laboratories.\cite{Danecek2011VCFtools,McKenna2010GATK,Li2009SAMtools}

Clinical genomics also established a culture of quality systems and validation. Early clinical NGS manuscripts emphasized analytical validation, reproducibility, and predefined reporting and interpretation policies before using results for patient management.\cite{Cottrell2014WUCaMP,Cheng2015JMolDiagn,Zehir2017NatMed} Multi-institutional biomarker-testing efforts (e.g., multiplexed driver testing consortia in lung cancer) further highlighted the operational requirements for translating molecular assays into clinical decision-making.\cite{Kris2014JAMA}

We argue that histopathology AI now occupies a similar position: strong technical progress, growing clinical interest, but incomplete consensus on standardized intermediate representations, QC metrics, and reporting conventions. Without these, external validation failures are difficult to interpret, benchmarking claims are hard to reproduce, and clinicians have limited visibility into model reliability.

\subsection*{What to Standardize for Pathology AI?}
GOLDMARK is motivated by the observation that pathology AI lacks NGS-like standardized intermediate files. In NGS, BAM and VCF files enable reproducible downstream analysis and consistent QC (e.g., depth, mapping quality, allele fraction) independent of the original sequencer and analysis code. By analogy, we propose and release a practical set of intermediate files for slide-level biomarker modeling: (i) \textbf{tile coordinate manifests} (a deterministic sampling manifest for each WSI), (ii) \textbf{per-slide feature tensors and embeddings with explicit encoder provenance}, (iii) \textbf{feature extraction metadata} to support QA/QC and cohort-level auditing, (iv) \textbf{trained slide-level weights and per-slide predictions}, and (v) \textbf{standardized metric tables and visualization outputs} aligned to those weights.

This work illustrates why these processes matter. Tile--feature integrity checks flag truncated or partial extraction that would otherwise silently bias model behavior, and embedding metadata provide signals for distribution shift monitoring and extraction-failure detection (Figure~\ref{fig:embedding_variance}). The public portal complements downloadable artifacts by providing consistent visualization and logging conventions (interactive ROC/PR curves, run summaries, and attention overlays) so that users can inspect the same evaluation objectives. The APIs available from the website allow programmatic downloads of intermediate files. 

\subsection*{MIL Baseline for Advancing Techniques}
We emphasize that GOLDMARK does \emph{not} claim gated-attention MIL is the final or dominant paradigm for computational pathology. We instead use a familiar and interpretable MIL head as a \textbf{reference implementation} that isolates the contribution of (i) cohort curation, (ii) clinically anchored labeling, (iii) standardized feature extraction, and (iv) external validation. This provides a stable ``common ground'' for the community: new aggregators, new training objectives, and new PFMs can be substituted while preserving the same intermediate artifacts, splits, and evaluation outputs.

Importantly, the proposed standards extend naturally to emerging approaches that move beyond fixed-feature MIL. Task-adapted pathology foundation models (TAPFMs) and related fine-tuning strategies incorporate task- and site-specific supervision into the encoder itself.\cite{TAPFM2025} EAGLE is an example of a task-adapted encoder built on top of a PFM backbone that can improve mutation prediction without requiring full end-to-end WSI training.\cite{EAGLE2025} EAGLE has also been independently evaluated as more generalizable than fixed-feature MIL baselines for \emph{EGFR} prediction under cross-cohort testing.\cite{rakaee_ancestry-associated_2026} Even as these approaches evolve, the same requirements for governed benchmarking remain: deterministic sampling, traceable intermediate representations (or their end-to-end equivalents), and standardized QC and reporting for fair cross-institutional comparisons.

\subsection*{Limitations and Future Work}
First, although we cover a broad set of clinically actionable tasks, many tumor--gene pairs remain challenging at current cohort sizes, particularly when morphologic correlates are subtle or heterogeneous. Second, MSKCC represents a single external institution; additional external sites are necessary to fully characterize generalization under broader practice variation. Third, we focused on a single interpretable MIL head to ground comparisons; future work should evaluate more recent aggregation strategies and end-to-end adaptation methods under the same standardized artifact and reporting framework.

Future releases will prioritize (i) expanding multi-institutional cohorts, (ii) extending standardized QC metrics (including slide-level and tile-level artifact scores) and reporting schemas for fine-tuned foundation models, and (iii) enabling governed community benchmarking workflows that preserve privacy and traceability while supporting reproducible comparisons.

\section*{Conclusions}
Computational pathology is entering a translational phase in which reproducibility, transparency, and cross-institutional robustness are prerequisites for clinical trust. GOLDMARK establishes a reference framework that separates dataset curation from model evaluation and introduces structured and downloadable intermediate files, quality-control metadata, and symmetric cross-dataset testing as core components of benchmarking. By standardizing these elements and releasing interoperable artifacts alongside baseline models, GOLDMARK shifts the focus from isolated model performance toward reproducible, comparable workflows. Such infrastructure-level standardization is essential for computational biomarkers to transition from promising research tools to dependable components of clinical oncology practice.

\section*{Data and Code Availability}
All GOLDMARK artifacts, including per-slide PFM feature tensors, tile coordinate manifests, slide-level MIL weights, and downloadable evaluation visualizations—are made available through \emph{www.artificialintelligencepathology.org} and mirrored in versioned archives. Code supporting curation, preprocessing, training, and internal/external inference is publicly available at \url{https://github.com/chadvanderbilt/GOLDMARK}. Pipelines can be re-capitulate from scratch with TCGA wholes slide images and the models can be run on users own slides by implementing feature extraction with public PFMs and downloding slide-level models from this study using GOLDMARK API.  The website provides the primary public interface for interactive exploration and artifact download, while the GitHub repository documents the underlying pipelines and reference implementation. Due to patient privacy constraints, raw MSKCC-IMPACT WSIs are not distributed publicly; the repository documents the required manifest schema and provides de-identified example headers to enable user replication under appropriate governance.

\section*{Funding}
Research funding was provided by a Cancer Center Support Grant from the NIH/NCI (grant number P30CA008748) and the Warren Alpert Foundation through the Warren Alpert Center for Digital and Computational Pathology at Memorial Sloan Kettering Cancer Center.

\section*{Competing Interests}
The authors declare no competing interests.

\nolinenumbers
\begingroup
\sloppy
\IfFileExists{vancouver.bst}{\bibliographystyle{vancouver}}{\bibliographystyle{unsrtnat}}
\bibliography{refs}
\endgroup
\linenumbers

\end{document}